\documentclass[runningheads]{llncs}
\usepackage{graphicx}
\usepackage{chngcntr}
\usepackage{multirow}
\usepackage[dvipsnames]{xcolor}
\usepackage{soul}

\counterwithout{table}{section}
\counterwithout{figure}{section}
\usepackage{hyperref}

\newcommand\blfootnote[1]{%
  \begingroup
  \renewcommand\thefootnote{}\footnote{#1}%
  \addtocounter{footnote}{-1}%
  \endgroup
}

\begin{document}

\title{A Feature Analysis for Multimodal News Retrieval} 

\author{Golsa Tahmasebzadeh\inst{1}\orcidID{0000-0003-1084-5552} \and
Sherzod Hakimov\inst{1}\orcidID{0000-0002-7421-6213} \and
Eric M\"uller-Budack\inst{1}\orcidID{0000-0002-6802-1241} \and
Ralph Ewerth\inst{1,2}\orcidID{0000-0003-0918-6297}}
\authorrunning{G. Tahmasebzadeh et al.}
%

\institute{TIB--Leibniz Information Centre for Science and Technology, Hannover, Germany \and
L3S Research Center, Leibniz University Hannover, Germany
\email{\{golsa.tahmasebzadeh, sherzod.hakimov, eric.mueller, ralph.ewerth\}@tib.eu}}

\maketitle              
\begin{abstract}
Content-based information retrieval is based on the information contained in documents rather than using metadata such as keywords. Most information retrieval methods are either based on text or image. In this paper, we investigate the usefulness of multimodal features for cross-lingual news search in various domains: politics, health, environment, sport, and finance. To this end, we consider five feature types for image and text and compare the performance of the retrieval system using different combinations. Experimental results show that retrieval results can be improved when considering both visual and textual information. In addition, it is observed that among textual features entity overlap outperforms word embeddings, while geolocation embeddings achieve better performance among visual features in the retrieval task. 

\keywords{Multimodal News Retrieval  \and Multimodal Features \and Computer Vision \and Natural Language Processing.}
\end{abstract}

\section{Introduction}
 The rapid growth of media content on the Web has led to a surge of intelligent technologies to organise them and satisfy users’ information needs. Multimodal information retrieval (MIR) is a branch of computer science that focuses on the identification of users' search needs and present them the most relevant resources considering information from different modalities. In today's Web era, one of the challenging aspects of retrieval is that information encoded in other formats than text are gaining importance, namely image, video, and audio data. Therefore, systems that utilize content from different modalities have received more and more attention in the research community in the last decade.\blfootnote{Copyright © 2020 for this paper by its authors. Use permitted under Creative Commons License Attribution 4.0 International (CC BY 4.0).}

In this paper, we analyse the impact of different features extracted from both text and image for information retrieval in the news domain. Prior work \cite{imagebert,websupervised} typically utilises state-of-the-art deep learning models for object recognition~\cite{vgg,resnet} or object detection \cite{fasterrcnn} to extract visual features. In contrast, we adopt three different visual descriptors including object, places and geolocation embeddings to cover images of different news domains. The difference of our approach with previous methods is that our visual descriptors are based on pre-trained deep learning architectures. For text, most state-of-the-art systems use Bidirectional Encoder Representations from Transformers (BERT) embeddings ~\cite{bert} to encode textual content. In addition, we consider another textual descriptor to analyse the overlap of entities mentioned in news articles. 
We focus on news domain and address five domains: politics, health, environment, sport, and finance, for both English and German language. We apply multimodal feature extraction on collected news articles that contain both image and text content. The ranking is calculated as a pair-wise similarity score between news articles based on either visual features, textual features, or their combinations. Given a query document, we compute the performance in terms of Average Precision (AP) of ranked documents. 

The main contribution of this paper is a comparison of different state-of-the-art feature descriptors for multimodal content, and how they affect the performance on information retrieval in the news domain.
Our analysis reveals that the combination of visual features and textual features performs better in comparison with each modality separately. Regarding textual features it is shown that entity overlap is an efficacious feature to describe news contents from different domains, while geolocation features from images perform better in different news domains when compared with object and places features. In general, the experiments show that simply taking the mean of multimodal features is already a good representative among all exclusive feature types.

The remainder of the paper is structured as follows. We discuss some related work on multimodal information retrieval in Section 2. Next in Section 3 we explain the collection of the dataset. In Section 4 the description of multimodal feature extraction for news article search is mentioned. We present the experimental results and discussions in Section 5. Finally, we conclude the paper with  findings for multimodal news retrieval using textual and visual features in Section 6.


\section{Related Work}
Initial methods for information retrieval are often based on only one modality and rely either on textual~\cite{deeptext} or on visual features~\cite{bagofvisual,cbi}.  Suarez et al.\cite{text_based_news} propose a method to collect related tweets to news articles by considering eight different search methods which are solely based on text such as: search by title, summary, content of text, bigram phrases and named entities to name but a few. More recently, Dai et al.~\cite{deeptext} explore the effect of BERT embeddings~\cite{bert} in Information Retrieval (IR) and show that enhancing word embeddings with additional knowledge from search logs produces a related search task in case of limited amount of labeled data. Saritha et al.~\cite{cbi} use Deep Belief Network (DBN) to extract visual features and report that the DBN generates a huge dataset for learning features and provides a good classification to handle the retrieval of relevant content.
 
The aforementioned approaches lack in representing content of a multimedia document since other modalities are not taken into account. To obviate this, multimodal-based methods were introduced~\cite{websupervised,imagebert}. 
Mithun et al.\cite{websupervised} learn an aligned image-text representation and update the joint representation using web images. On the other hand, Qi et al.\cite{imagebert} train a multitask model on four different tasks to model the linguistic information and visual content. Mithun et al.\cite{websupervised} in addition to visual and textual features leverage web images with noisy tags to overcome the limited labeled data. However, Qi et al.\cite{imagebert} collect a Large-scale weAk-supervised Image-Text (LAIT) from the Web to enhance pre-training and further fine-tune the model using public datasets in a multi-stage format. Both state-of-the-art multimodal approaches are focused on increasing training data to improve the performance, but do not incorporate different visual and textual descriptors to represent image and text more comprehensively. A different approach is proposed by Vo et al.\cite{composed} to retrieve images where query is an image along with a description given by user. They combine image and text through Compositional Learning where core idea is that a complex concept can be developed by combing multiple simple concepts or attributes \cite{cl}.
Crossmodal consistency is another approach which is useful in news retrieval~\cite{mm_consistency}. M{\"{u}}ller{-}Budack et al. \cite{mm_consistency} proposed a multimodal approach to quantify cross-modal entity coherence between image and text by gathering visual evidence from the Web using named entity linking.


Besides visual and textual features, modalities other than image and text are also of interest to improve the performance of a MIR system. For instance Dang{-}Nguyen et al.\cite{geocbi} apply geolocation coordinates as additional information. They adopt support vector machine and apply bag-of-words as visual feature vector, and user-generated tags as textual features. Then, the model is trained to assign the optimal weights for each descriptor. They report that this extra information significantly improves the performance.

Inspired by the above mentioned methods, we combine different visual and textual descriptors and show the impact of each descriptor in different news domains.


\section{Dataset}
In order to collect an appropriate dataset for the envisioned feature analysis, we extracted news articles from five news domains: politics, health, environment, sport, and finance. For each domain, we manually selected recent or impactful news events, for instance, \textit{Brexit} for politics and \textit{Coronavirus} for the health domain. We gathered a maximum of 20 news articles for 25 events in English and German using the EventRegistry\footnote{\url{http://eventregistry.org/}} API (Application Programmer's Interface). In total, we obtained 348 English and 263 German articles. Then two experts manually verified if the crawled news articles match the queried event. More information on the extracted dataset is provided in Table~\ref{table:dataset}. Each extracted news article contains a title, body text, and an image. 

\begin{table}[t]
\centering
\caption{Number of retrieved news articles for each domain and language}
\begin{tabular}{lcc}
\textbf{Domain} & \textbf{English} & \textbf{German} \\
\hline
Politics              & 94                & 35               \\
Environment           & 96                & 70               \\
Health                & 82                & 54               \\
Sport                & 43                & 61               \\
Finance               & 33                & 43              \\
\hline
Total               & 348                & 263              
\end{tabular}
\label{table:dataset}
\end{table}

\section{Methodology}
In Section 4.1, we explain how the extraction of multimodal features is performed using pre-trained deep learning approaches. Then, we describe the computation of pair-wise similarities between news articles to do the retrieval task.

\subsection{Multimodal Features}
\begin{figure}[t]
\includegraphics[width=\textwidth]{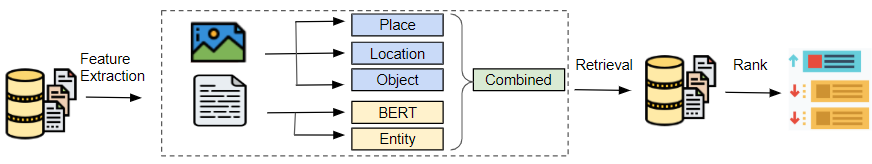}
\caption{Multimodal Information Retrieval (MIR) system for news articles by considering multimodal features from text and image together.
} \label{fig1}
\label{fig:feature_extraction}
\end{figure}

Prior work often utilises features from a single modality be it either text or visual content. Considering the variety of images and textual content used in news, we aim to analyse the effects of multimodal features for news information retrieval. Using different types of features to represent an image or text is crucial in an information retrieval system, specially in news retrieval. There are various categories of articles such as sport, environment and politics, each of which requires distinct descriptors to represent the content of the news. For instance, in environmental images places and geolocation are more important than objects; in sport different types of visual features such as objects, places, and geolocation are necessary to represent all aspects of an image. We use the embeddings of pre-trained convolutional neural networks from state-of-the-art computer vision for object detection, place recognition, and geolocation estimation as visual features. Entity vectors and word embeddings serve as textual features.
The process of multimodal feature extraction is shown in Figure~\ref{fig:feature_extraction}. Each news article contains a title, body text, and an image.

\subsubsection{Visual Features}
To extract visual features from images, three different visual descriptors are adopted: objects, places, and geolocations. Since news articles are usually from different domains, their corresponding images have distinct types of visual information. To extract features pre-trained deep learning models for different tasks are applied to extract rich feature vectors from the last fully-connected layer of the respective convolutional neural network. Please note that we do not take the predictions from these models but the weights (that lead to model predictions). Regarding the models as explained below, this is the layer before the last softmax activation.

\begin{itemize}
    \item Object recognition: We use the ResNet-50~\cite{resnet} model pre-trained on the ImageNet dataset~\cite{imagenet}, where the task is to recognise 1000 distinct objects in images such as car, person, etc. The dimension of the resulting feature vector for an image is 2048.
    \item Place recognition: We use the ResNet~\cite{resnet_v2} model pre-trained on the Places365 dataset~\cite{365places_dataset}, where the task is to recognize 365 distinct places such as beach, stadium, street etc. The dimension of the resulting feature vector for an image is 2048.
    \item Geolocation recognition: We use the model~\cite{geolocation} based on ResNet101~\cite{resnet_v2} pre-trained on a subset of the Yahoo Flickr Creative Commons 100 Million dataset (YFCC100M)~\cite{yfcc}. The subset, which includes around five million geo-tagged images, was introduced for the MediaEval Placing Task 2016 (MP-16)~\cite{mediaeval}. This model is aimed at predicting the geolocation of an image. The dimension of the resulting feature vector for an image is 2048.
\end{itemize}

Each image of an article was fed into the models described above and three 2048-dimensional vectors for objects, places, and geolocation were extracted.

\subsubsection{Textual Features}

Textual features are extensively used in information retrieval systems since most of the context in news are provided in textual format. Therefore, we consider two different features to retrieve relevant documents for comparing the textual content. The first feature type comprises named entities in a given news article, while the second type of features are word embeddings representing the text. We assume that similar events mention similar entities, as well as similar events being described with similar words. Thus, the overlap of entities and similarity of word embeddings between articles are important features for information retrieval.

First, we explain how to extract named entities from news articles. As mentioned above, we consider English and German news articles that cover various events from five domains. We use spaCy~\cite{spacy} to extract named entities and Wikifier~\cite{wikifier} to link those named entities to Wikipedia pages, since these tools support both languages. First, we extract named entities and their corresponding spans in a text using spaCy. Then, we use Wikifier to extract named entities, their spans and additionally the links to Wikipedia pages and PageRank score for each detected entity.

We combine the outputs from both systems by considering the spans of extracted named entities where both spaCy and Wikifier agree on. We select the linked entity from Wikifier with the highest PageRank score with the aim of disambiguation. Finally, we collect named entities with their links to Wikipedia pages for both English and German news articles.

\begin{itemize}
    \item Entity vectors:  As mentioned above, we collect all extracted named entities in order to convert each news article into a vector representation. Each news article is converted into an entity vector representation, where an entry in the vector is set to 1, if the entity (related to the entry) appears in a document, otherwise it is set to 0. In total, the news articles contained 5,195 and 1,991 entities for English and German, respectively.
    \item Word embeddings: We use BERT~\cite{bert} embeddings to extract word vectors for all sentences in text, since such word embeddings take into account the contextual surrounding of words. Since text content of news articles is long, we use a sliding window approach by selecting 1500 characters at a time and extracting word vectors from BERT. We use the last layer of the output, where a 768-dimensional vector is assigned to each token. The word vectors for each token are then averaged to obtain a single vector representing the given span from text. We continue the process until all tokens are processed. The resulting vectors represent the whole news text in terms of word embeddings using BERT.
\end{itemize}

\subsection{Multimodal News Retrieval}
\label{subsec:mir}
In this section, we describe the retrieval task performed in this paper. After collecting news articles, the five feature embeddings are computed for image and text as explained above. The retrieval system is essentially returning a list of relevant documents for a given query. In our case, the query is a news article and the task id to retrieve news articles of the same event based on uni-modal and multimodal similarity measures.

We compute a pair-wise similarity between news articles using cosine similarity between selected vectors depending on the modality. The similarity of news articles is computed separately for each language. Each news article is treated as a query and the remaining ones as a reference to compute cosine similarity. The remaining news articles are ranked by their similarity score in regard to the selected query. As described above, we consider five different features from image and text. We average the similarity scores from each feature when the modalities are merged for the retrieval task.

 
The evaluation of the performance is based on average precision score using Eq.\ref{eqn:avgP}. In this equation \textit{P} stands for Precision, \textit{R} stands for Recall and \textit{n} defines the $n^{th}$ threshold. We use this measure because it combines recall and precision in different thresholds for ranked retrieval results, thus, better represents the overall performance. In other words, for one information need, the average precision is mean of the precision scores regarding different thresholds after each relevant document is retrieved.
\begin{equation}
\label{eqn:avgP}
Average Precision = {\sum_{n}({R_n-R_{n-1}}){P_n}}
\end{equation}
In this paper AP is calculated by looking at the ranked list of other news articles whether they are relevant or not. For instance, we pick an article from the event \textit{Brexit} and rank the rest of news articles by the similarities to the chosen one. The objective of the retrieval task is to rank the remaining news articles in the same event higher than others.

\section{Experiments}
In this section, we discuss evaluation results to measure the performance of the proposed multimodal information retrieval system. To better demonstrate the performance, evaluation is done in different configurations by considering information from different modalities as follows:

\begin{itemize}
 \item Only textual features
 \item Only visual features
 \item Visual and textual features
\end{itemize}

In all of the configurations Average Precision is used as a performance measure as explained in Section~\ref{subsec:mir}.

\subsection{Evaluation Results}

We evaluate the performance of each modality separately and in combination. We provide the evaluation of the proposed system for each selected event in Table~\ref{table:eng} and Table~\ref{table:deu} for English and German news articles respectively. In these tables, each row presents average precision of the corresponding event using a single feature or a combination of features. The combination of features is done by averaging the similarity scores from the corresponding features. We computed the performance for each feature, combination of features from the same modality, and combination of both modalities. The best performing features are highlighted in bold for each event in Table~\ref{table:eng} and Table~\ref{table:deu}.

As shown in Table~\ref{table:eng} and Table~\ref{table:deu} regarding textual features, the first three columns show average precision using only textual features including: BERT embeddings (B), entity overlap (E) and mean of both features ($\overline{T}$) respectively. Among individual textual features, for English, entity overlap achieves the best performance since it outperforms in five events, while BERT embeddings outperform in only two events, as highlighted in the Table~\ref{table:eng}. Similarly, for German, feature entity overlap achieves the best performance since it outperforms in five events, while BERT embeddings outperforms in only one event as highlighted in the Table~\ref{table:deu}. Regarding combination of textual features for English it outperforms each individual textual feature by achieving the best average precision in six events, and for the German news by outperforming in five events it equals to entity overlap performance.
 
Regarding visual features, the next four columns show results for three visual features including: objects (O), places (P), geolocation (L), and combined ($\overline{V}$). Regarding individual visual features in comparison with all other eight features for English, in only three events, and for German in five events either of features: objects, places and geolocation, outperform the rest. For English, individual visual features in comparison with each other, have similar performance. For German, geolocation has better performance than the others since it outperforms in three events, while objects and places both outperform in only one event in total. As mentioned above, mean approach is considered as combination of features where the similarities of combined features are averaged. For English, in none of the events mean of visual features ($\overline{V}$) outperforms the rest of features, whereas for German it outperforms in four events in total as presented in Table \ref{table:deu}. 

We consider the same combination regarding all the five feature types for both visual and textual by averaging the similarity scores. As shown in both tables, out of 25 events, regarding English, the three different combinations including: mean of all features (V+T), mean of visual ($\overline{V}$) and mean of textual ($\overline{T}$) features in comparison with each other outperform in eleven, zero and six of events respectively.
For the German news the mentioned performances are eight, four and five respectively. Thus, it is evident that for both languages combination of visual and textual outperforms each individual feature (B, E, O, P, L) and the combination of visual ($\overline{V}$) and textual ($\overline{T}$) features.

The results presented in Table~\ref{table:events} are average precision scores for five domains: Politics, Sport, Health, Environment, and Finance. It is observed that for English the mean of all features (T+V) outperforms in three out of five events, which are \textit{Environment}, \textit{Health} and \textit{Sport}. Similar pattern is observed for German where the domains are: \textit{Politics}, \textit{Environment} and \textit{Finance}. Therefore, for both languages for three out of five news domains the combination of multimodal features resulted in a better performance for the information retrieval task.



\begin{table}[]
\centering
\fontsize{8}{14}\selectfont
\setlength\tabcolsep{3pt}
\caption{Investigation of textual (B: BERT, E: Entity overlap, $\overline{T}$: Mean of textual features), visual features (O: Object, P: Places, L: Geolocation, $\overline{V}$: Mean of visual features) and combination of textual and visual (T+V) features for English news articles covering different domains and events regarding average precision. Note that the values are multiplied by 100. The highest score for each event among other features is highlighted in bold font.}
\begin{tabular}{|l|l|c|c|c|c|c|c|c|c|c|c|c|}
\hline
                           & & \multicolumn{3}{c|}{\textbf{Textual~(T)}} & \multicolumn{4}{c|}{\textbf{Visual~(V)}} & \multicolumn{1}{c|}{\textbf{T+V}} \\ \hline
                             & \textbf{Event} & B    & E    & $\overline{T}$        & O   & P  & L   & $\overline{V}$   & Mean                           \\ \hline 
\hline

\hline \multirow{6}{*}{\rotatebox{90}{\textbf{Politics}}} & 2016 United States presidential election& 14& 39& 35& 23& 35& 30& 31& \textbf{42} \\
\cline{2-10}  & Impeachment of Donald Trump& 12& 68& 62& 55& 53& 59& 60& \textbf{72} \\
\cline{2-10}  & European Union–Turkey relations& 12& \textbf{41}& 38& 14& 9& 16& 14& 28 \\
\cline{2-10}  & War in Donbass& 3& \textbf{55}& 47& 3& 7& 12& 11& 20 \\
\cline{2-10}  & Brexit& 8& \textbf{73}& 68& 19& 16& 21& 21& 44 \\
\cline{2-10}  & Cyprus–Turkey maritime zones dispute& 49& 82& \textbf{87}& 66& 60& 59& 61& 80 \\
\hline
\hline \multirow{7}{*}{\rotatebox{90}{\textbf{Environment}}} & Global warming& 10& 6& 9& 9& \textbf{11}& 8& 9& \textbf{11} \\
\cline{2-10}  & Water scarcity& \textbf{17}& 8& 13& 10& 15& 11& 13& \textbf{17} \\
\cline{2-10}  & 2019–20 Australian bushfire season& 2& 19& 20& \textbf{36}& 17& 12& 23& 33 \\
\cline{2-10}  & Indonesian tsunami& 7& 66& 61& 34& 44& 48& 50& \textbf{67} \\
\cline{2-10}  & Water scarcity in Africa& 12& 19& 22& 11& 13& 15& 15& \textbf{23} \\
\cline{2-10}  & 2018 California wildfires& 8& 60& 55& 42& 35& 37& 44& \textbf{65} \\
\cline{2-10}  & Palm oil production in Indonesia& 19& 18& 25& 23& 43& 48& 45& \textbf{50} \\
\hline
\hline \multirow{3}{*}{\rotatebox{90}{\textbf{Finance}}} & Financial crisis of 2007–08& 9& 9& \textbf{11}& 4& 7& 7& 6& 7 \\
\cline{2-10}  & Greek government-debt crisis& 11& \textbf{63}& 62& 8& 6& 15& 12& 47 \\
\cline{2-10}  & Volkswagen emissions scandal& 7& \textbf{76}& 70& 30& 36& 47& 46& 71 \\
\hline
\hline \multirow{5}{*}{\rotatebox{90}{\textbf{Health}}} & Coronavirus& 50& 65& \textbf{68}& 18& 17& 31& 29& 66 \\
\cline{2-10}  & Ebola virus disease& 11& 23& 24& 21& 21& 34& 34& \textbf{37} \\
\cline{2-10}  & Zika fever& 15& 27& 29& 28& 32& 42& 41& \textbf{48} \\
\cline{2-10}  & Avian influenza& 13& 13& 16& 21& 21& 25& 26& \textbf{32} \\
\cline{2-10}  & Swine influenza& 36& 28& \textbf{40}& 16& 19& 20& 22& 35 \\
\hline
\hline \multirow{4}{*}{\rotatebox{90}{\textbf{Sport}}} & 2016 Summer Olympics& 12& 32& 28& 37& 36& \textbf{62}& 57& 60 \\
\cline{2-10}  & 2018 FIFA World Cup& 8& 23& \textbf{24}& 13& 14& 14& 17& 22 \\
\cline{2-10}  & 2020 Summer Olympics& 12& 53& \textbf{58}& 7& 10& 11& 10& 29 \\
\cline{2-10}  & 2022 FIFA World Cup& \textbf{37}& 13& 16& 12& 8& 8& 11& 16 \\
\hline

\end{tabular}
\label{table:eng}
\end{table}

\begin{table}[]
\centering
\fontsize{8}{14}\selectfont
\setlength\tabcolsep{3pt}
\caption{Investigation of textual (B: BERT, E: Entity overlap, $\overline{T}$: Mean of textual features), visual features (O: Object, P: Places, L: Geolocation, $\overline{V}$: Mean of visual features) and combination of textual and visual (T+V) features for German news articles covering different domains and events regarding average precision. Note that the values are multiplied by 100. The highest score for each event among other features is highlighted in bold font.}
\begin{tabular}{|l|l|c|c|c|c|c|c|c|c|c|c|c|}
\hline
                           & & \multicolumn{3}{c|}{\textbf{Textual~(T)}} & \multicolumn{4}{c|}{\textbf{Visual~(V)}} & \multicolumn{1}{c|}{\textbf{T+V}} \\ \hline
                             & \textbf{Event} & B    & E    & $\overline{T}$        & O   & P  & L   & $\overline{V}$   & Mean                           \\ \hline \hline

\hline \multirow{6}{*}{\rotatebox{90}{\textbf{Politics}}} & 2016 United States presidential election& \textbf{6}& 2& 2& 2& 3& 2& 3& 3 \\
\cline{2-10}  & Impeachment of Donald Trump& 2& 36& 29& 24& 33& 33& 30& \textbf{38} \\
\cline{2-10}  & European Union–Turkey relations& 3& 3& 3& 34& \textbf{37}& 34& 35& 35 \\
\cline{2-10}  & War in Donbass& 4& \textbf{77}& 65& 13& 10& 10& 13& 28 \\
\cline{2-10}  & Brexit& 3& 22& 22& 12& 17& 14& 15& \textbf{28} \\
\cline{2-10}  & Cyprus–Turkey maritime zones dispute& 2& 18& 9& 53& 55& 63& \textbf{65}& 56 \\
\hline
\hline \multirow{7}{*}{\rotatebox{90}{\textbf{Environment}}} & Global warming& 19& 26& 29& 21& 22& 20& 23& \textbf{34} \\
\cline{2-10}  & Water scarcity& 7& 12& 11& \textbf{22}& 21& 16& 21& 17 \\
\cline{2-10}  & 2019–20 Australian bushfire season& 3& 69& 62& 53& 54& 47& 53& \textbf{72} \\
\cline{2-10}  & Indonesian tsunami& 4& 3& 4& 3& 5& \textbf{28}& 9& 7 \\
\cline{2-10}  & Water scarcity in Africa& 41& 42& \textbf{43}& 7& 10& 24& 16& 38 \\
\cline{2-10}  & 2018 California wildfires& 2& 16& 8& 23& 20& \textbf{25}& \textbf{25}& 23 \\
\cline{2-10}  & Palm oil production in Indonesia& 14& 28& 32& 23& 31& 39& 37& \textbf{43} \\
\hline
\hline \multirow{3}{*}{\rotatebox{90}{\textbf{Finance}}} & Financial crisis of 2007–08& 11& \textbf{38}& 29& 18& 21& 30& 25& 33 \\
\cline{2-10}  & Greek government-debt crisis& 5& 10& 9& 6& 9& \textbf{14}& 11& 13 \\
\cline{2-10}  & Volkswagen emissions scandal& 8& 19& 16& 27& 36& 34& 37& \textbf{38} \\
\hline
\hline \multirow{5}{*}{\rotatebox{90}{\textbf{Health}}} & Coronavirus& 2& \textbf{57}& 41& 13& 18& 16& 16& 30 \\
\cline{2-10}  & Ebola virus disease& 10& \textbf{49}& 42& 20& 28& 35& 34& \textbf{49} \\
\cline{2-10}  & Zika fever& 5& 3& 5& 17& 16& 19& \textbf{20}& 18 \\
\cline{2-10}  & Avian influenza& 27& 41& \textbf{43}& 23& 19& 30& 27& 37 \\
\cline{2-10}  & Swine influenza& 3& 14& \textbf{19}& 2& 5& 4& 4& 12 \\
\hline
\hline \multirow{4}{*}{\rotatebox{90}{\textbf{Sport}}} & 2016 Summer Olympics& 12& 60& \textbf{61}& 18& 20& 27& 26& \textbf{61} \\
\cline{2-10}  & 2018 FIFA World Cup& 6& 11& 8& 24& 24& 24& \textbf{25}& 21 \\
\cline{2-10}  & 2020 Summer Olympics& 6& \textbf{45}& 39& 11& 15& 17& 15& 36 \\
\cline{2-10}  & 2022 FIFA World Cup& 21& 37& \textbf{40}& 10& 10& 13& 11& 22 \\
\hline


\end{tabular}
\label{table:deu}
\end{table}

\begin{table}[t]
\centering
\fontsize{8}{14}\selectfont
\setlength\tabcolsep{3pt}
\caption{Comparison of multimodal features regarding average precision scores for different news domains in German and English news articles. T: textual features, V: visual features, T+V: textual and visual features combined. The highest score for each event category is highlighted in bold font.}
\begin{tabular}{|c|c|c|c|c|c|c|}

\hline
                     & \multicolumn{3}{c|}{\textbf{English}} & \multicolumn{3}{c|}{\textbf{German}} \\ \hline
                  \textbf{Domain}   & $\overline{T}$       & $\overline{V}$     & T+V    & $\overline{T}$    & $\overline{V}$     & T+V     \\ \hline


Politics&  \textbf{55}&  32&  47&  21&  26&  \textbf{30}\\ \hline
Environment&  28&  28&  \textbf{37}&  26&  25&  \textbf{33}\\ \hline
Finance&  \textbf{47}&  21&  41&  17&  23&  \textbf{27}\\ \hline
Health&  34&  30&  \textbf{43}&  \textbf{29}&  19&  28\\ \hline
Sport&  \textbf{31}&  23&  \textbf{31}&  \textbf{36}&  19&  34\\ \hline

\end{tabular}
\label{table:events}
\end{table}

\subsection{Discussion}
As mentioned earlier, Table~\ref{table:events} shows the impact of different features in different domains, and Table~\ref{table:eng} and Table~\ref{table:deu} show the evaluation results for each event associated with each domain. In this section we further study the numbers reported in the tables and discuss the impact of features in different domains.

To compare visual and textual features together, as presented in Table~\ref{table:events}, for English, in all the categories textual features are better descriptors than visual features, except for \textit{Environment} where both features have equal average precision score. For German, three categories including \textit{Sport}, \textit{Environment} and \textit{Health} are the ones that fit this condition. The reason that in English news textual features are better than visual features in more categories than German is that entity overlap in English in total obtains a better performance than German. In more detail, the named entity extraction tool, spaCy, extracted more entities in English than in German. Thus, in German news retrieval, for some queries the obtained entity overlap similarities with the reference articles are zero. In these cases we set the similarity scores to very small random number.

Regarding combination of all features, in English news even though visual features are not better than textual features, they helped textual features improve the overall performance for domains such as \textit{Environment} and \textit{Health} (see T+V column in Table~\ref{table:events}). On the other hand, for \textit{Politics} and \textit{Finance} textual features outperform either visual and combined features. One reason is that the content of images in these domains are not noticeable in terms of places, geolocation or objects. The other reason is the richness of text in comparison with images. Since these two domains include very specific events such as \textit{Volks\-wagen emissions scandal} and \textit{Greek government debt crisis}, due to specific entities existing in their texts, entity overlap outperforms the other four remaining feature types including all visual features (see column T+V in Table~\ref{table:eng}). Therefore, the experiments show that there is a need for additional visual descriptors to better represent the visual content. For instance, face detectors that distinguish depicted persons in images might be helpful, since usually there are popular people in images of these news domains.

Regarding textual features individually, as presented in Table~\ref{table:events}, in English news, \textit{Politics} is the one that achieved the highest performance using only textual features for which the reason is that events such as \textit{Cyprus-Turkey maritime zones dispute} report higher in comparison with events in other categories using entity overlap as a textual descriptor. However, \textit{Environment} is the one with the least average precision using only textual features. The reason is that events such as \textit{Water scarcity} or \textit{Global warming} are broad topics where the chances of having a big entity overlap is low. In addition, it is observable from the results in Table~\ref{table:eng} and Table~\ref{table:deu} that BERT embeddings in most cases did not yield any improvements over the other features. Conversely, the entity overlap in most cases outperforms all the other individual feature types. Thus, it is worth to mention that in news retrieval systems instead of comparing the whole text it is better to focus on named entities mentioned in text. 

From visual point of view, for German in most events geolocation and combined features outperform the other two visual features objects and places, and for English individual visual features in total outperform the combined visual features. As mentioned in Section 5.1, visual features do not outperform either textual or combination of all features (T+V). One possible reason for the low performance of visual descriptors might be that the models that are used in this research are trained on domains other than news. Therefore, they are not able to extract useful visual clues from news images. Nevertheless, they have a good impact in improving the average precision, in the retrieval task, when combined with textual features as presented in Table~\ref{table:eng}, Table~\ref{table:deu} and Table~\ref{table:events}.

\section{Conclusion}

In this paper, we have proposed a feature analysis for multimodal news retrieval, considering and representing both image and text content in news articles. To this end, we have investigated the impact of three visual descriptors (objects, places, and geolocation) as well as two textual descriptors (entity overlap and text similarity using BERT embeddings). 

We evaluated the approach on 25 events extracted from five news domains. Experiments show that multimodal (combination of visual and textual) features outperform individual visual and textual features. Furthermore, we showed that the textual feature of entity overlap performs better than BERT embeddings for both English and German news articles. We observed that in some domains additional visual descriptors such as face detectors might help on top of the existing visuals features.

In future work, we intend to train a supervised model that learns to assign different importance weights for the available features values. Another approach could be to increase the set of features to better represent images of different news domains to improve the overall performance when combined with textual features. Besides, extending the dataset by including more news domains and other languages for more in depth experiments is another future direction.

\section*{Acknowledgements}

This project has received funding from the European Union’s Horizon 2020 research and innovation programme under the Marie Skłodowska-Curie grant agreement no 812997.

\bibliographystyle{splncs04}
\bibliography{references}

\end{document}